\documentclass{article}

     \PassOptionsToPackage{numbers, compress}{natbib}

\usepackage[preprint]{neurips_2021}

\usepackage[utf8]{inputenc} 
\usepackage[T1]{fontenc}    
\usepackage{hyperref}       
\usepackage{url}            
\usepackage{booktabs}       
\usepackage{amsfonts}       
\usepackage{nicefrac}       
\usepackage{microtype}      
\usepackage{xcolor}         
\usepackage{makecell}
\usepackage{graphicx}
\usepackage{amsmath}
\usepackage{listings}
\title{Self-supervised Feature Enhancement: Applying Internal Pretext Task to Supervised Learning}

\author{%
	\small Yuhang Yang\thanks{University of Electronic Science and Technology of China}\\
	\texttt\small{yuhang.y@std.uestc.edu.cn} \\
	\And
	\small Zilin Ding$^{*}$ \thanks{Equally contributed}\\
	\texttt\small{dingzilin@std.uestc.edu.cn} \\
	\And
	\small Xuan Cheng$^{*}$ \\
	\texttt \small{cs\_xuancheng@std.uestc.edu.cn} \\
	\AND
	\small Xiaomin Wang$^{*}$\thanks{Second corresponding author}  \\
	\texttt\small{xmwang@uestc.edu.cn} \\
	\And
	\small Ming Liu$^{*}$\thanks{First corresponding author} \\
	\texttt\small{csmliu@uestc.edu.cn} \\
}

\begin{document}
	
\maketitle

\begin{abstract}
	Traditional self-supervised learning requires CNNs using external pretext tasks (i.e., image- or video-based tasks) to encode high-level semantic visual representations. In this paper, we show that feature transformations within CNNs can also be regarded as supervisory signals to construct the self-supervised task, called \emph{internal pretext task}. And such a task can be applied for the enhancement of supervised learning. Specifically, we first transform the internal feature maps by discarding different channels, and then define an additional internal pretext task to identify the discarded channels. CNNs are trained to predict the joint labels generated by the combination of self-supervised labels and original labels. By doing so, we let CNNs know which channels are missing while classifying in the hope to mine richer feature information. Extensive experiments show that our approach is effective on various models and datasets. And it's worth noting that we only incur negligible computational overhead. Furthermore, our approach can also be compatible with other methods to get better results.
\end{abstract}

\section{Introduction}
\label{Introduction}

In recent years, self-supervised learning \cite{doersch2015unsupervised} has achieved remarkable success in various visual feature learning tasks, such as image classification \cite{noroozi2016unsupervised}, semantic segmentation \cite{long2015fully}, object detection \cite{girshick2014rich}. In the absence of manual annotations, this approach uses only unlabeled data to obtain pseudo labels automatically through a proposed pretext task, and captures semantic features via predicting the pseudo labels. The proposed pretext task is constructed by \emph{external} signals, and usually image- or video-based, such as image geometric transformation recognition \cite{gidaris2018unsupervised}, image colorization \cite{zhang2016colorful}, video prediction \cite{srivastava2015unsupervised}, etc. To solve these tasks, CNNs can learn visual feature representations of images, which are then further used as a pre-training model for downstream visual tasks.

Even though self-supervision was initially proposed for the lack of large-scale labeled datasets, self-supervised technique has recently been applied to other fields due to its simplicity and effectiveness. $S^4L$ uses self-supervision to improve the performance of semi-supervised learning \cite{zhai2019s4l}. SS-GAN trains Generative Adversarial Networks (GANs) with self-supervised methods \cite{chen2019self}. And SLA proposes a framework that combines self-supervised techniques with supervised learning \cite{lee2020self}. But this external image-based method, that could be viewed as expand the dataset several times, entails multiple times the training overhead of the baseline. This motivates us to exploit the possibility of internal pretext task.

Our work is based on the following assumption: different channels correspond to different aspects of semantic features (see Figure \ref{cam}). If the network can distinguish which channels are missing, it will have a deeper understanding of the whole feature map. It's just like the blind man touching the elephant: if he can know a part of the elephant through each touch, then with enough touches he not only knows the different parts of the elephant, but also knows more about the elephant as a whole. We term this kind of self-supervised task using internal network signals \emph{internal pretext task}.

Specifically, we propose a self-supervised task by recognizing feature transformations, and apply it for the enhancement of supervised learning, which only incurs negligible time cost and memory overhead. We first define the feature transformations: divide the channels of the feature map into several groups, and discard a group of channels each time to get a new feature map. Then we concatenate all the feature maps and transmit them into the rest of the network. Each feature map corresponds to a joint label, that is, the combination of the original label and self supervised label, and the model is trained to predict the joint labels. Moreover, in order to prevent the overhead increase caused by the generated feature maps being transferred to the next convolutional layer, we further introduce a two-classifier structure.

Our contributions are:
\begin{itemize}
	\item We first propose and prove that feature transformations within CNNs can be regarded as supervisory signals to construct the self-supervised task, and term this self-supervised task using internal network signals \emph{internal pretext task}.
	\item We propose a generic framework that combines our internal pretext task with supervised learning, which incurs negligible computational overhead and significantly improves the classification accuracy.
	\item We exhaustively evaluate our method in various networks(e.g., ResNet-18, ResNet-110, PyramidNet-110-270) and various datasets (e.g., CIFAR-100, tiny-ImageNet, CUB200), which demonstrated the wide applicability, versatility and compatibility of our method. 
\end{itemize}

\section{Related Work}
\label{Related Work}

Self-supervision is a generic learning framework that was proposed to learn visual features from large-scale unlabeled images or videos without any human annotations. In self-supervised learning, CNNs generate self-supervised labels through a proposed pretext task, and improve the ability to extract features for downstream tasks. Due to its unique advantage of training model with only unlabeled data, Self-supervision has been widely applied in various computer vision tasks. In this paper, we apply self-supervised technique into a supervised learning framework to enhance the classification of fully labeled images. And in this part, we first make an incomplete summary of self-supervised learning, and then introduce several methods of applying self-supervised techniques to other domains including supervised learning.

\paragraph{Self-supervised learning}

\cite{doersch2015unsupervised} trains a CNN model by artificially constructing a pretext task which is defined as predicting the relative position between two random patches in the same image. Based on this idea, a jigsaw puzzle was designed in 2016 as a pretext task: training the model to put nine scrambled blocks back to their original positions \cite{noroozi2016unsupervised}. Follow-up papers \cite{ahsan2019video,kim2018learning,wei2019iterative} formulate different jigsaw puzzles to learn visual representations. In addition to the above patch-based approaches, \cite{gidaris2018unsupervised} designs a pretext task to predict the rotation angle of images. Furthermore, \cite{misra2016shuffle} is proposed as a temporal context-based method in which the pretext task is formulated to determine whether the frame sequence in the video is arranged in the correct temporal order. And \cite{lee2017unsupervised} trains a model to learn features by sorting frame sequences.

\paragraph{Self-supervised techniques in other domains}

SS-GAN proposes a self-supervised Generative Adversarial Network (GAN) \cite{chen2019self}. It uses a self-supervised method based on image rotations to generate labels, and then achieves the goal by an auxiliary loss. $S^4L$ is a framework of semi-supervised loss caused by self-supervised learning objectives \cite{zhai2019s4l}. The authors believe that self-supervised learning techniques can greatly benefit from a small number of labeled examples. And SLA applies self-supervised methods to supervised learning \cite{lee2020self}. In the framework of SLA, the original classification task and the self-supervised task are treated as a joint task and a CNN model is trained to learn the joint task. SLA combines the generated labels with the original labels, and then solves the joint task by predicting joint labels of images.

\section{Methods}
\label{Methods}

\begin{figure}[htbp]
	\centering
	\includegraphics[width=1.0\linewidth]{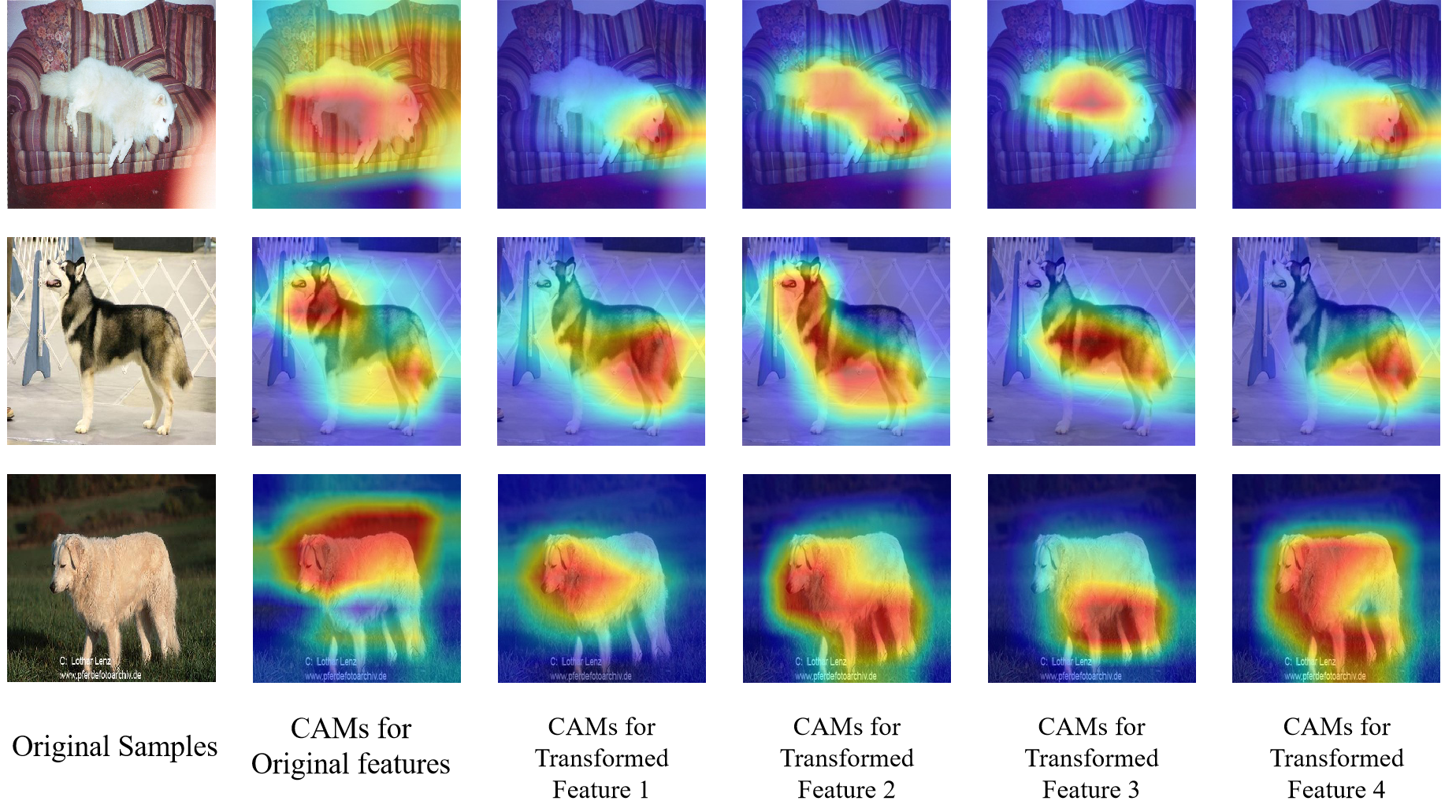}
	\caption{Class activation maps (CAMs) \cite{Zhou2016Learning} for ResNet-18 model trained on Stanford Dogs. Each column represents the visualization results of the same network in different images. The second column shows the results of training the baseline model, and the third to sixth columns show the results of the models discarding 4 different groups of channels in the last layer, which means that only part of the features of the last layer can participate in the network training. The visualization results show that discarding different channels makes the network pay attention to different aspects of features.}
	\label{cam}
\end{figure}

\paragraph{Motivation.} SLA enhances supervised learning by a pretext task of recognizing image transformations, e.g., image rotations. However, SLA demands to generate multiple transformed images for each image, which is equivalent to expanding the dataset several times, and thus inevitably incurs a great deal of time consumption and memory overhead. This inspires us to design a self-supervised task that uses the internal signals of the network. To this end, we consider transforming the feature maps by discarding different channels, and define a self-supervised task to recognize the discarded channels. Intuitively, different channels correspond to different aspects of semantic features. To prove this, we train four models discarding four different groups of channels in the last layer, which means that only part of the features can participate in the network training. We demonstrate their class activation maps (CAMs) \cite{Zhou2016Learning} in Figure \ref{cam}. The visualization results show that features which discard different channels focus on different semantic information, which indicates that this feature transformation can be regarded as a supervisory signal to construct the self-supervised task. And we argue that if the network can distinguish which channels are missing, it will have a deeper understanding of the whole feature map.

\paragraph{Notation.} Let $ x \in \mathbb{R}^{W \times H \times C} $ with label $ y \in \{1,...,N\} $ be the input where $ N $ is the number of classes, $H$, $W$, and $C$ denote height, width, and the number of channels respectively. Let $\mathcal{L}_{\mbox{{\tiny CE}}}$ denote the cross-entropy loss function, $D_{\mbox{{\tiny KL}}}$ denote the KL-divergence loss function, $ F(\cdot|\,\theta)$ denote a CNN model for fully labeled datasets, where $ \theta $ are the learnable parameters of model $ F(\cdot) $. Specifically, we define a set of $ k $ transformations $ T=\{t_{j}\}^{k}_{j=1} $, where $ t_{j} $ is the operator that applies to the feature map $ f=F(x|\,\theta) $. And $ \tilde{f}_{j}=t_{j}(f) $ is a transformed sample by $ t_{j} $. Let $K=k+1$ represent the number of total feature maps (Including $k$ transformed feature maps and 1 original feature map). 

\paragraph{Transformation.} To propose a new type of internal pretext task, we first define the transformations $T$ of features. The key idea of $T$ is to discard specific channels of the feature map $f$. For a given feature map of size $\mbox{height}\times\mbox{width}\times n_{\mbox{{\tiny channels}}}$, we randomly divide all the channels of the feature map into $ k $ groups. Each group contains $n_{\mbox{{\tiny channels}}}/k$ channels. And for each transformation $ t_{j} $, we generate a binary mask $M_{j}$ of size $\mbox{height}\times\mbox{width}\times n_{\mbox{{\tiny channels}}}$. If a channel is chosen to be discarded, the corresponding $\mbox{height}\times\mbox{width}$ values in $M_{j}$  are all set to 0, and other values in $M_{j}$ are set to 1. By multiplying the original feature map $f$ with $M_{j}$, we get a new feature map $\tilde{f}_{j}$. Namely, we define the transformation operation $t_{j}$ as:
\begin{equation}\label{Trans1}
	\tilde{f}_{j}= t_{j}(f)= f \odot M_{j}
\end{equation}
where each transformed feature map ${f}_{j}$ is an incomplete version of the original feature map $f$. It's worth noting that their missing parts do not cross each other, and the summation of the missing parts is just the original feature map, which is:
\begin{equation}\label{Trans2}
	f= \sum_{j=1}^{k}(f-\tilde{f}_{j})
\end{equation}
\paragraph{Self-supervised feature enhancement.} Now we move on to the whole scheme of our method, with further discussion about how we define internal pretext task and how we enforce it in enhancing the supervised learning. We take the images $x$ with only original labels $y$ as the input, and train a model to obtain the corresponding feature map $f$ of each image. Then, through the pre-defined transformations $T$, i.e., discarding different channels of the feature map, $k$ groups of feature maps are generated, and self-supervised labels $j$ are assigned for each feature map. Next, we concatenate all the feature maps (including a original one) and transmit them into the rest of the network, and finally into a joint classifier $\sigma(\cdot;\mu)$. Each feature map corresponds to a joint label ($y$,$j$), that is, the combination of the original label $y$ and the self supervised label $j$ , and the network is trained to predict the joint labels. 

\begin{figure}[htbp]
	\centering
	\includegraphics[width=1.0\linewidth]{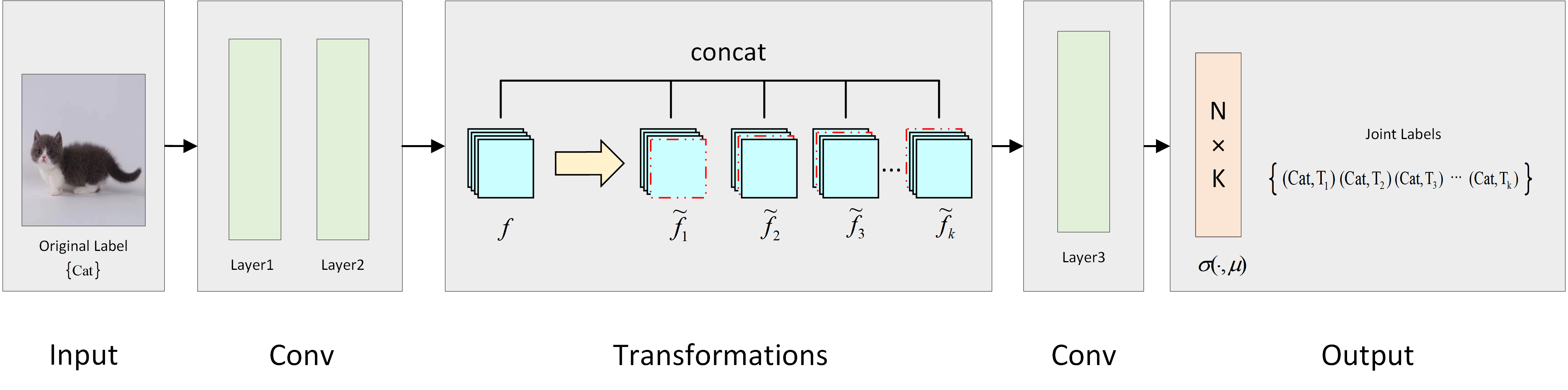}
	\caption{An specific training pipeline of our proposed network architecture after second convolutional layer (taking ResNet as an example). After an input image is passed into the network and through several convolutional layers, the feature map $ f $ is obtained. Then the channels of $f$ are randomly divided into $ k $ groups, and one group of channels is discarded each time to get $ \tilde{f}_{1},\tilde{f}_{2},\cdots,\tilde{f}_{k} $ . Finally, All of them are concatenated together and transmitted to a joint classifier.}
	\label{architecture}
\end{figure}

Figure \ref{architecture} shows the specific training pipeline of introducing internal pretext task after the second convolutional layer. For example, when training on CIFAR-100 \cite{krizhevsky2009learning} (100 labels), if we divide the total channels into 4 groups, and discard a group of channels as a transformation each time, then we can get 5 groups of feature maps (including the original one), so the model learns the joint probability distribution on all possible combinations, i.e., 500 labels. 

Since the model uses a joint classifier $\sigma(\cdot;\mu)$, the joint probability of the final output can be written as:
\begin{equation}\label{joint probability}
	P(i,j|\,\tilde{f})=\sigma_{ij}(\tilde{f};\mu)=\frac{\exp(\mu_{ij}^{\top}\,\tilde{f})}{\sum_{k,l}\exp(\mu_{kl}^{\top}\,\tilde{f})}
\end{equation}
Therefore, given a set of $ M $ training images $ D=\{x_{i}\}^{M}_{i=1} $, the training objective that the CNN model must learn to solve is:
\begin{equation}\label{training objective}
	\min_{\theta}\frac{1}{M}\sum_{i=1}^{M}loss(x_{i},\theta)
\end{equation}
where the loss function $ loss(\cdot) $ is defined as:
\begin{equation}\label{loss}
	loss(x,\theta)=\frac{1}{K}\sum_{j=1}^{K}\mathcal{L}_{\mbox{{\tiny CE}}}(\sigma(\tilde{f};\mu),(y,j))
\end{equation}

\begin{figure}[htbp]
	\centering
	\includegraphics[width=1.0\linewidth]{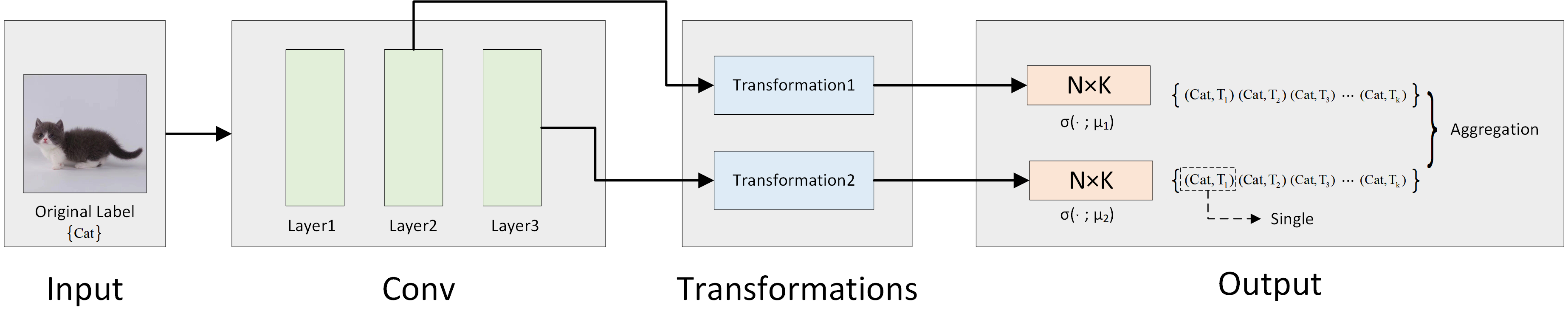}
	\caption{An overview of our proposed two-classifier structure. The features of the last two layers are transformed, and transmitted to two classifiers respectively.we use the original feature of the last layer for single inference, and for aggregated inference, we aggregate the probability of all $ 2 \times K $ features. }
	\label{two-classifier structure}
\end{figure}

\paragraph{Two-classifier structure.} Since the newly generated feature maps will continue to be transferred into the rest convolutional layers, which increases the computational cost, we further propose a two-classifier structure in the hope of cost reduction. As shown in Figure \ref{two-classifier structure}, feature maps are transformed after the last two layers, and transmitted to two joint classifiers $\sigma(\cdot;\mu_{1})$ and $\sigma(\cdot;\mu_{2})$ respectively. We usually avoid adding a classifier after the first layer because the features of the first layer are not trained enough to obtain high-level semantic visual representations. And then we train the network by optimizing the summation of their losses. The total loss function is defined as:
\begin{equation}\label{two classifier loss}
	\begin{aligned}
		\mathcal{L}_{\mbox{{\tiny 2FC}}}&=\mathcal{L}_{\mbox{{\tiny penultimate}}}+\beta\cdot\mathcal{L}_{\mbox{{\tiny last}}}\\&=\frac{1}{K}\sum_{j=1}^{K}\mathcal{L}_{\mbox{{\tiny CE}}}(\sigma(\tilde{f};\mu_{1}),(y,j))+\beta\cdot\frac{1}{K}\sum_{j=1}^{K}\mathcal{L}_{\mbox{{\tiny CE}}}(\sigma(\tilde{f};\mu_{2}),(y,j))
	\end{aligned}
\end{equation}
where $\beta$ is a hyperparameter in the range of $[0,1]$  to balance the impact of the classifier after the penultimate layer on the network. By introducing this two-classifier structure, although we increase the network parameters (an additional classifier is introduced), we can prevent features from being transferred to the next layer, which can significantly reduce the overhead and further improve the performance. We use this two-classifier structure for all experiments unless otherwise noted.

\section{Experiments}
\label{Experiments}
In this section we conduct an extensive evaluation of our approach on the most commonly used image datasets, including three benchmark image classification datasets CIFAR-10/100 \cite{krizhevsky2009learning}, tiny-ImageNet\footnote{https://tiny-imagenet.herokuapp.com/} and three fine-grained image datasets CUB200 \cite{wah2011the}, Stanford Dogs \cite{KhoslaYaoJayadevaprakashFeiFei_FGVC2011}, Stanford Cars \cite{KrauseStarkDengFei-Fei_3DRR2013}, as well as on multiple models, such as ResNet \cite{he2016deep}, SeNet \cite{hu2018squeeze}, DenseNet \cite{huang2017densely}, Wide ResNet \cite{zagoruyko2016wide}, PyramidNet \cite{han2017deep}. Furthermore, we also compare our approach with SLA \cite{lee2020self}, a recent proposed self-supervised method which is used in supervised learning, and combine it with SLA and other regularization methods. In all experiments, we refer baselines which use only random cropping and flipping for data augmentation as "Baseline". We provide implementation details of our experiments in the supplementary material.

\subsection{Evaluation Method}
\label{Evaluation Method}
Although the final joint classifier predicts the probability of $ N \times K $ labels for all $ K $ feature maps, we only care about the softmax probability of the original feature $ f $ on the original $ N $ labels, and then compute the accuracy of single inference. So we use the single softmax probability $P(i|\,f,j=1)$ to predict its label, which is as follows:
\begin{equation}\label{single probability}
	P(i|f,j=1)=\frac{\exp(\mu_{i1}^{\top}f)}{\sum_{k=1}^{N}\exp(\mu_{k1}^{\top}f)}
\end{equation}

We also introduce aggregated inference in our framework. Given a feature map, we can only consider the original $N$ labels because we already know which transformation is applied. Therefore, we use the conditional probability $ P(i|\,\tilde{f}_{j},j)$ to predict its original label. Furthermore, we aggregate the probability of all $ K $ feature maps to obtain comprehensive performance. And if there are two joint classifiers in the model, we aggregate the probability of all $ 2 \times K $ feature maps (see Figure \ref{two-classifier structure}. The aggregated probability can be written as:
\begin{equation}\label{aggregated probability}
	P_{\mbox{{\tiny agg}}}(i|x)=\frac{\exp(z_{i})}{\sum_{k=1}^{N}\exp(z_{k})}
\end{equation}
where $ z_{i}=\frac{1}{K}\sum_{j=1}^{K}\mu_{ij}^{\top}\tilde{f}_{j} $. And in order to restore the basic network structure in the test phase, we introduce self-distillation \cite{hinton2015distilling} to transfer the knowledge of the joint classifier $\sigma(\cdot;\mu)$ to another single classifier $\sigma(\cdot;v)$. In this way, images can be classified using only a basic network model, and the accuracy obtained is the most standard. To this end, we add a single loss $ \mathcal{L}_{\mbox{{\tiny CE}}}(\sigma(f;v),y) $ to train the single classifier, and use a KL-loss $ D_{\mbox{{\tiny KL}}}(P_{\mbox{{\tiny agg}}}(\cdot|x)||\sigma(f;v)) $  to make it learn from the joint classifier. After training, we only use the single classifier for inference without aggregation. In our experiments, we denote our three inference schemes, i.e., the single inference, the aggregated inference, and the inference of self-distillation as SFE-SI, SFE-AG and SFE-SD.

\subsection{Image Classification}

\paragraph{Classification on CIFAR-100.}
We empirically verify the effectiveness of our proposed method on benchmark image dataset CIFAR-100. To this end, we adopt ResNet-110, ResNet-164, SE-ResNet-110, DenseNet-100-12, Wide ResNet-28-10 and PyramidNet-110-270 as the baselines to evaluate our framework. Experiments show that our framework has significant improvement on all six models compared to the baseline. In particular, our method improves the top-1 accuracy of the cross-entropy loss from 74.21\% to 76.84\% of ResNet-110 under the CIFAR-100 dataset (see Table \ref{cifar100 table}). The results show that our method has a strong generalization to networks, and it can have a good performance whether on ResNet or other large networks such as PyramidNet.

And we notice that in our experiments, the aggregated inference does not improve the classification accuracy significantly compared to the single inference, and sometimes even leads to performance degradation. The reason is that part of the semantic visual representations are discarded when we transform the features, and if the discarded semantic visual representations happen to be an important part of the target to be identified, it may cause the network to get an incorrect classification result, which will play a negative role in aggregation. Experiments also show that self-distillation can well transfer the knowledge of the joint classifier to the single classifier.

\begin{table}[b]
	\caption{Validation accuracy (\%) on CIFAR-100 and tiny-ImageNet using various CNN architectures. SFE\,-\,SI, SFE\,-\,AG and SFE\,-\,SD indicate the single inference, the aggregated inference and the inference of self-distillation, respectively. (The best accuracy is indicated as bold).}
	\label{cifar100 table}
	\centering
	\begin{tabular}{lllll}
		\toprule[1.3pt]
		\midrule
		Model  & Baseline &\quad SFE\,-\,SI         	 &\quad SFE\,-\,AG        &\quad SFE\,-\,SD  \\
		\midrule
		ResNet-110 & 74.21{\small \,$\pm$\,0.35} &\quad 76.80{\small \,$\pm$\,0.22} &\quad \textbf{76.84{\small \,$\pm$\,0.25}} & \quad 76.24{\small \,$\pm$\,0.05}\\
		ResNet-164 & 75.10{\small \,$\pm$\,0.15} &\quad 77.33{\small \,$\pm$\,0.12}&\quad 77.53{\small \,$\pm$\,0.13}&\quad \textbf{77.68{\small \,$\pm$\,0.13}} \\
		SE-ResNet-110 & 76.38{\small \,$\pm$\,0.12} &\quad 77.50{\small \,$\pm$\,0.01}&\quad 77.59{\small \,$\pm$\,0.07}&\quad \textbf{77.75{\small \,$\pm$\,0.11}}\\
		DenseNet  & 77.60{\small \,$\pm$\,0.13}&\quad \textbf{78.41{\small \,$\pm$\,0.01}}&\quad 78.30{\small \,$\pm$\,0.05} &\quad 78.34{\small \,$\pm$\,0.05} \\
		WRN-28-10  & 79.30{\small \,$\pm$\,0.21} &\quad \textbf{80.71{\small \,$\pm$\,0.10}}&\quad 80.66{\small \,$\pm$\,0.06}&\quad 80.70{\small \,$\pm$\,0.01}\\
		PyramidNet & 81.54{\small \,$\pm$\,0.31} &\quad \textbf{82.82{\small \,$\pm$\,0.04}}&\quad 82.63{\small \,$\pm$\,0.06}&\quad 81.98{\small \,$\pm$\,0.03}\\
		
		\bottomrule[1.3pt]
	\end{tabular}
\end{table}

\paragraph{Classification on CIFAR-10 and tiny-ImageNet.}
Furthermore, our method has also wide applicability, versatility, and compatibility for datasets. It can also achieve good results on other image datasets such as CIFAR-10 and tiny-ImageNet. As shown in Table \ref{cifar10 and tinyimagenet table}, our framework achieves general improvements on CIFAR-10 and tiny-ImageNet, whether using the single inference, the aggregated inference, or the inference of self-distillation. Especially, we achieve the accuracy of training tiny-ImageNet on ResNet-110 to 63.79\%.

\begin{table}
	\caption{Validation accuracy (\%) on CIFAR-10, tiny-ImageNet using ResNet-110. SFE\,-\,SI, SFE\,-\,AG and SFE\,-\,SD indicate the single inference, the aggregated inference and the inference of self-distillation, respectively.  The best accuracy is indicated as bold.}
	\label{cifar10 and tinyimagenet table}
	\centering
	\begin{tabular}{lllll}
		\toprule[1.3pt]
		\midrule
		Dataset  & Baseline & \quad SFE\,-\,SI         	 & \quad SFE\,-\,AG        & \quad SFE\,-\,SD  \\
		\midrule
		CIFAR-10 	   & 94.59{\small \,$\pm$\,0.05}&\quad \textbf{95.06{\small \,$\pm$\,0.06}}&\quad 94.99{\small \,$\pm$\,0.08}&\quad 94.95{\small \,$\pm$\,0.03}\\
		tiny-ImageNet & 61.94{\small \,$\pm$\,0.06}&\quad \textbf{63.79{\small \,$\pm$\,0.34}}&\quad 63.47{\small \,$\pm$\,0.28}&\quad 63.04{\small \,$\pm$\,0.02} \\
		\bottomrule[1.3pt]
	\end{tabular}
\end{table}

\paragraph{Comparison with SLA.}
We compare our method with SLA, a recent proposed self-supervised method which is used in supervised learning. SLA expands data from the image level, while our framework only increases the number of features after convolutional layers, so the incremental cost is far less than that of SLA. Table \ref{comparison with sla} compares the time consumption and memory overhead of our method with SLA. As can be seen from table \ref{comparison with sla}, compared with the basic network, our method only increases the time consumption by 24.1\% and the GPU memory by 21.3\%, while SLA increases the cost by almost 3 times. Moreover, our method is comparable to SLA in the ability of the single inference. We cost much less than SLA and the accuracy is only 0.45\% lower than SLA.

\begin{table}
	\caption{The cost and single-inference accuracy of our method compared to baseline and SLA for training with CIFAR-100 on ResNet-110. All experiments run on the same GPU.}
	\label{comparison with sla}
	\centering
	\begin{tabular}{llll}
		\toprule[1.3pt]
		\midrule
		Method & \makecell[c]{GPU memory\\(MiB)} & \makecell[c]{Training time\\(sec/iter)}&  \makecell[c]{Single Accuracy\\(\%)}  \\ 
		\midrule
		baseline &\qquad 2846 {\small($+$\,0.0\%)}              &\quad 0.178 {\small($+$\,0.0\%)}       &\quad  74.21{\small \,$\pm$\,0.35}          \\ 
		SFE     &\qquad 3455 {\small($+$\,21.3\%)}                 &\quad 0.221 {\small($+$\,24.1\%)}       &\quad  76.80{\small \,$\pm$\,0.22}          \\ 
		SLA      &\qquad 9718 {\small($+$\,241.3\%)}                 &\quad 0.693 {\small($+$\,289.3\%)}      &\quad  77.25{\small \,$\pm$\,0.03}          \\ 
		\bottomrule[1.3pt]
	\end{tabular}
\end{table}

\paragraph{Combination with SLA and regularization methods.}
Furthermore, our method can also be compatible with SLA and other regularization methods to get better results. As shown in Table \ref{combination with sla}, we apply our method with SLA and existing regularization techniques, Mixup \cite{zhang2017mixup},Cutout \cite{devries2017improved}. In order to be unified with SLA, we only use the output of the last classifier for aggregated inference in this experiment. The experimental results show that our method consistently reduces the classification errors. As a result, it achieves 19.03\% error rate when training CIFAR-100 on ResNet-110. These results also demonstrate the compatibility of our method.

\begin{table}
	\caption{Classification error rates (\%) of SLA and other regularization methods with our method on CIFAR-100. The minimum error rate is indicated as bold.}
	\label{combination with sla}
	\centering
	\begin{tabular}{ll}
		\toprule[1.3pt]
		\midrule
		Method   			&\qquad\qquad       Error Rate   \\ 
		\midrule
		ResNet110				& \qquad\qquad\quad      25.79  \\ 
		+ Cutout				& \qquad\qquad\quad      24.49  \\ 
		+ Cutout + SFE (ours)			& \qquad\qquad\quad      22.59  \\ 
		+ Mixup     			& \qquad\qquad\quad      22.30  \\ 
		+ Mixup + SFE (ours)      	& \qquad\qquad\quad      21.10  \\ 
		+ SLA					& \qquad\qquad\quad      19.71  \\ 
		+ SLA + SFE	(ours)		& \qquad\qquad\quad    	 19.53  \\ 
		+ Cutout + SLA			& \qquad\qquad\quad    	 19.43  \\ 
		+ Cutout + SLA + SFE (ours)	& \qquad\qquad\quad    	 19.03  \\ 
		\bottomrule[1.3pt]
	\end{tabular}
\end{table}

\subsection{Fine-grained Image Classification}

The fine-grained image classification aims to recognize similar subcategories of objects under the same basic-level category. The difference between fine-grained recognition and general category recognition is that fine-grained subcategories usually share the same parts, which can only be distinguished by the subtle differences in texture and color characteristics of these parts. We adopt ResNet-18 as the baseline, and use CUB200, Stanford Dogs, Stanford Cars, three commonly used fine-grained datasets to verify that our proposed method also achieves good results on fine-grained classification tasks. As shown in Table \ref{fine-grained datasets table}, our improvement on fine-grained datasets is remarkable, especially on Stanford Dogs, which improves by 6.62 percentage points (63.58\% vs 56.96\%).

\begin{table}
	\caption{Validation accuracy (\%) on three fine-grained datasets CUB-200, Stanford Dogs and Stanford Cars using ResNet-18. SFE\,-\,SI, SFE\,-\,AG and SFE\,-\,SD indicate the single inference, the aggregated inference and the inference of self-distillation, respectively.  The best accuracy is indicated as bold.}
	\label{fine-grained datasets table}
	\centering
	\begin{tabular}{lllll}
		\toprule[1.3pt]
		\midrule
		Dataset  & Baseline & \quad SFE\,-\,SI         	 & \quad SFE\,-\,AG        & \quad SFE\,-\,SD  \\
		\midrule
		CUB200  	   & 69.17{\small \,$\pm$\,0.32} & \quad \textbf{73.01{\small \,$\pm$\,0.36}} & \quad 72.26{\small \,$\pm$\,0.20} & \quad 71.54{\small \,$\pm$\,0.38}\\
		Stanford Dogs & 56.96{\small \,$\pm$\,0.31} & \quad 63.43{\small \,$\pm$\,0.01}  & \quad \textbf{63.58{\small \,$\pm$\,0.01}} &  \quad 63.55{\small \,$\pm$\,0.02}  \\
		Stanford Cars & 80.01{\small \,$\pm$\,0.56} & \quad 85.18{\small \,$\pm$\,0.24}  & \quad \textbf{85.33{\small \,$\pm$\,0.27}} & \quad 84.11{\small \,$\pm$\,0.01}\\
		\bottomrule[1.3pt]
	\end{tabular}
\end{table}

\subsection{Ablation Study}

\paragraph{Exploring the pattern of feature transformation.}

In Table \ref{discussion of K and layers} we explore how the performance of the network depends on the number of Transformations $ K $, and layers where the features are transformed. We use CIFAR-100 to experiment on ResNet-110. For that purpose we randomly divide all channels into 4, 8, 16, 32 groups, i.e., $ K = 5, 9, 17, 33 $. We do not consider the case of $ K > 33 $ because of the high computational cost. In the meantime the situation of introducing internal pretext task after the first layer, the second layer and the third layer is considered respectively. In this experiment if the transformation occurs after the first or the second layer, the generated features will be transferred to the rest of the network, and finally to the joint classifier. 

We observe that the earlier internal pretext task is proposed, the better the learned features are, because they get more training. For the second and third layers, within a certain range, as the self-supervised task become more difficult, we achieve better image classification performance. While for the first layer, the best performance is achieved when $ K = 5 $. We speculate that the more difficult task can excavate deeper feature information after the second and third layers, but after the first layer, the original classification task has not been trained enough, so the difficult self-supervised task will interfere with the original task, resulting in performance degradation.

In general, we can get the best accuracy by using $ K = 5 $ transformations after the first layer, 3.06 percentage points higher than baseline (77.27\% vs 74.21\%). And we can get a balance between the cost increase and performance improvement with transformations $K = 9$ after the second layer. And to further reduce the computational overhead, we propose a multi-classifier structure (a two-classifier structure is shown in Figure \ref{two-classifier structure}). By the way, we have also considered the cases of artificial channel division, partial channel division and one channel divided into multiple groups. Finally we think that it is more reasonable to divide all channels randomly.

\begin{table}
	\caption{The effect of the number of Transformations $ K $ and layers where the features are transformed on network performance. SI and AG indicate the single inference and the aggregated inference of the single classifier structure respectively.}
	\label{discussion of K and layers}
	\centering
	\resizebox{\textwidth}{15mm}{
		\begin{tabular}{lllllll}
			\toprule[1.3pt]
			\midrule
			& \multicolumn{2}{c}{Layer1} & \multicolumn{2}{c}{Layer2} & \multicolumn{2}{c}{Layer3} \\ \cmidrule(r){2-3}\cmidrule(r){4-5}\cmidrule(r){6-7}
			&\qquad SI         	 & \qquad AG          &\qquad SI           &\qquad AG          &\qquad SI           &\qquad AG          \\
			\midrule
			K = 1 + 4  & \textbf{77.02{\small \,$\pm$\,0.13}} 		 & \textbf{77.27{\small \,$\pm$\,0.14} } 	   & 75.20{\small \,$\pm$\,0.02}  	  & 75.34{\small \,$\pm$\,0.03}      & 75.10{\small \,$\pm$\,0.06}       & 75.16{\small \,$\pm$\,0.07}       \\
			K = 1 + 8  & 76.79{\small \,$\pm$\,0.10}          & 76.90{\small \,$\pm$\,0.16}     & 75.85{\small \,$\pm$\,0.07}         & 76.23{\small \,$\pm$\,0.10}      & 75.56{\small \,$\pm$\,0.18}          & 75.58{\small \,$\pm$\,0.17}         \\
			K = 1 + 16 & 76.21{\small \,$\pm$\,0.11}       & 76.22{\small \,$\pm$\,0.17}      & 76.14{\small \,$\pm$\,0.04}           & 76.27{\small \,$\pm$\,0.07}         &  75.57{\small \,$\pm$\,0.07}         &  75.60{\small \,$\pm$\,0.24}        \\
			K = 1 + 32 & 76.87{\small \,$\pm$\,0.08}        & 76.89{\small \,$\pm$\,0.13}      & \textbf{76.42{\small \,$\pm$\,0.08}}        & \textbf{76.36{\small \,$\pm$\,0.07}}       & \textbf{75.79{\small \,$\pm$\,0.08}}   	   &  \textbf{75.83{\small \,$\pm$\,0.08}}     \\ 
			\bottomrule[1.3pt]
	\end{tabular}}
\end{table}

\paragraph{Evaluation of the multi-classifier structures.}\label{Evaluation of the multi-classifier structure}
We explore different design choices of multi-classifier structure, these choices are classified into two groups. The first group discusses the number of layers that use classifiers. ResNet-110 contains three layers, we use classifiers after the last layers, last two layers and whole three layers, respectively. As shown in Table \ref{discussion of classifiers}, we can get the best accuracy when we use classifiers after the last two layers. We speculate that the reason why the performance is degraded when we add the classifier of the first layer is that features of the first layer are not trained enough to encode high-level semantic visual representations. 

The second group foucuses on whether the internal pretext task is proposed or not. To further determine the benefits of our proposed pretext task, we compare the results using only multiple classifiers with the results using multiple classifiers and the pretext task. As shown in Table \ref{discussion of classifiers}, our proposed self-supervised task works well no matter how many classifiers are used, and we get the best result when we use two classifiers. Therefore, in our framework, we use a two-classifier structure and combine it with our proposed internal pretext task.

\begin{table}
	\caption{The effect of different design choices of multi-classifier structure on network performance. SI and AG indicate the single inference and the aggregated inference of the single classifier structure respectively.}
	\label{discussion of classifiers}
	\centering
	\begin{tabular}{lllll}
		\toprule[1.3pt]
		\midrule
		& \multicolumn{2}{c}{ResNet110} & \multicolumn{2}{c}{ResNet110\,+\,SFE} \\ \cmidrule(r){2-3}\cmidrule(r){4-5}
		&\qquad SI         	 &\quad \qquad AG          &\quad \qquad SI           &\quad \qquad AG            \\
		\midrule
		1 Classifier & 74.21{\small \,$\pm$\,0.35} & \quad 74.21{\small \,$\pm$\,0.35} & \quad 75.79{\small \,$\pm$\,0.08} & \quad\textbf{75.83{\small \,$\pm$\,0.08}}               \\ 
		2 Classifiers & 75.34{\small \,$\pm$\,0.25} & \quad 75.99{\small \,$\pm$\,0.07} &\quad 76.80{\small \,$\pm$\,0.22} &\quad\textbf{76.84{\small \,$\pm$\,0.25}}          \\ 
		3 Classifiers & 74.88{\small \,$\pm$\,0.28} & \quad 74.75{\small \,$\pm$\,0.11} &\quad \textbf{75.98{\small \,$\pm$\,0.03}}&\quad 75.88{\small \,$\pm$\,0.09} \\
		\bottomrule[1.3pt]
	\end{tabular}
\end{table}

\paragraph{Discussion on the setting of the hyperparameter $\beta$.}

\begin{figure}[htbp]
	\centering
	\includegraphics[width=0.7\linewidth]{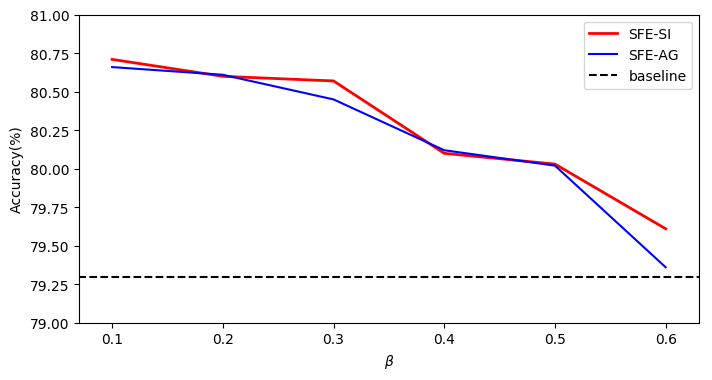}
	\caption{Validation accuracy under different settings of $\beta$. We train CIFAR-100 on Wide ResNet-28-10, and report the accuracy of the single inference (blue line) and the aggregated inference (yellow line). Baseline represents the result of the basic model which is nothing to do with $\beta$ (red line).}
	\label{beta-curve}
\end{figure}

As is known to us, the quality of the learned features after the penultimate layer is not as good as that of the last layer. In order to balance the impact of the learned features after the penultimate layer and the last layer on the network, we introduce a hyperparameter $\beta$ to reduce the weight of the loss of the penultimate layer.  As shown in Figure \ref{beta-curve}, for example, if we train CIFAR-100 on Wide ResNet-28-10, the network performance fluctuates with $\beta$, and generally decline with the increase of $\beta$, so we need to set a small weight $\beta$ to suppress the impact of the output features after the penultimate layer. We suspect that the quality of the features after the penultimate layer may be much worse than that of the last layer. When $\beta$ increases too much, the features after the penultimate layer will have a negative impact on the training of the network.

\section{Conclusion}
\label{Conclusion}
We have introduced a novel self-supervised pretext task that uses internal signals of the network, termed internal pretext task. Then we have proposed a generic framework that applies this internal pretext task for the enhancement of supervised learning. Extensive experiments show that our approach is effective on various models and datasets. More broadly, we hope it could stimulate discussion in the community regarding this very important, but largely neglected perspective: internal signals of the network can also be applied to construct self-supervised tasks.
\medskip
\small{
	\bibliographystyle{plain}
	\bibliography{ref}
}

\newpage
\appendix

\section{Implementation Details}
\label{Implementation Details}
We choose ResNet-110 as the backbone model for CIFA10/100 and tiny-ImageNet, and ResNet-18 as the backbone model for the fine-grained datasets CUB200, Stanford Dogs and Stanford Cars unless otherwise stated, and use SGD, momentum of 0.9 and weight decay of $ 10^{-4} $. We set the learning rate to 0.5 for PyramidNet-110-270 while for other models, we set it to 0.1. For all datasets, we train for 300 epochs and decay the learning rate by a factor of 0.1 at epoch 150 and 225. For CIFAR-10/100, we train with training batch size of 128 (except 64 on DenseNet) and test batch size of 100. For tiny-ImageNet, we train with training batch size of 256 and test batch size of 100. And for the fine-grained datasets, we train with training batch size of 16 and test batch size of 64. For CUB200, we set the input size of the images to $448 \times 448$, while for Stanford Dogs and Stanford Cars, the input size is $224 \times 224$. We define $K=9$ transformations unless otherwise stated. We set $\beta=0.1$ for Wide ResNet, PyramidNet, DenseNet, $\beta=0.5$ for ResNet-110, ResNet-164, SE-ResNet-110, and $\beta=1.0$ for ResNet-18. We report the average accuracy of four trials for all experiments unless otherwise noted. When combining with other methods, we use publicly available codes and run them on our network and settings.

\section{Implementation with Pytorch}
\label{Implementation with Pytorch}
We implement our code using Pytorch on 4$\times$Tesla M40 GPUs. Each GPU contains 24 GB of memory. One of the advantages of our method is simple to implement on various models. We first define the feature transformation with a code segment (see \emph{Listing 1}). For a given feature map of size $\mbox{height}\times\mbox{width}\times n_{\mbox{{\tiny channels}}}$, we generate $k$ binary masks to discard channels. Then we respectively multiply the original feature and $k$ masks to obtain $k$ transformed features, and concatenate them together to get a joint feature map. Next, we apply the transformation code and implement self-supervised feature enhancement on various networks (an example on ResNet is shown in \emph{Listing 2}). In this way, we finally complete the framework of SFE by expanding the labels $k$ times during the network training. We will upload the complete code to Github later. We believe that the simplicity of our method could lead to the wide applicability for various applications.

\lstset{
	basicstyle          =   \sffamily,
	keywordstyle        =   \bfseries,
	commentstyle        =   \rmfamily\itshape,
	stringstyle         =   \ttfamily,
	flexiblecolumns, 
	showspaces          =   false,
	numberstyle         =   \tiny\ttfamily,
	showstringspaces    =   false,
	captionpos          =   t,
	language        =   Python,
	basicstyle      =   \ttfamily,
	numberstyle     =   \ttfamily,
	keywordstyle    =   \color{blue},
	keywordstyle    =   [2]\color{teal},
	stringstyle     =   \color{magenta},
	commentstyle    =   \color{red}\ttfamily,
	tabsize		=	4, 
}
\begin{lstlisting}[title={\emph{Listing 1.} Class definition of feature transformation.},label=listing1]
class Transaction:
	def __init__(self, size, n):
		c, h, w = size
		drop = torch.zeros((h, w)).cuda()
		seed_list = random.sample(range(0, c), c)
		self.n = n
		self.mask_list = []
		for i in range(n):
			mask = torch.ones(size).cuda()
			mask[seed_list[i * c // n:(i + 1) * c // n:1], :, :] = drop
			self.mask_list.append(mask)

	def trans1(self, x):
		b, c, h, w = x.size()
		mask = torch.ones((b, c, h, w)).cuda()
		x1 = x.cuda() * mask
		for i in range(self.n):
		y = x1.cuda() * self.mask_list[i].unsqueeze(0).repeat(b, 1, 1, 1).cuda()
			x = torch.cat((x, y), 1)
		x = x.view(-1, *x1.shape[1:])
		return x
\end{lstlisting}

\begin{lstlisting}[title={\emph{Listing 2.} Self-supervised feature enhancement using ResNet.},label=listing2]
class ResNet(nn.Module):
	def __init__(self, depth, num_transform=9, num_classes=100, block_name='Bottleneck'):
		...
		self.l2_transaction = Transaction((32 * block.expansion, 16, 16), num_transform - 1)
		self.l3_transaction = Transaction((64 * block.expansion, 8, 8), num_transform - 1)
		...
	
	def forward(self, x):
		x = self.conv1(x)
		x = self.bn1(x)
		x = self.relu(x)
		x = self.layer1(x)
		x = self.layer2(x)
		x2 = self.l2_transaction.trans1(x)
		x2 = self.l2_avgpool(x2)
		x2 = x2.view(x2.size(0), -1)
		x2 = self.l2_fc(x2)
		x = self.layer3(x)
		x3 = self.l3_transaction.trans1(x)
		x3 = self.l3_avgpool(x3)
		x3 = x3.view(x3.size(0), -1)
		x3 = self.l3_fc(x3)
		return x2, x3
\end{lstlisting}

\section{Explanation of Self-distillation}
\label{Explanation of Self-distillation}
\begin{figure}[htbp]
	\centering
	\includegraphics[width=0.4\linewidth]{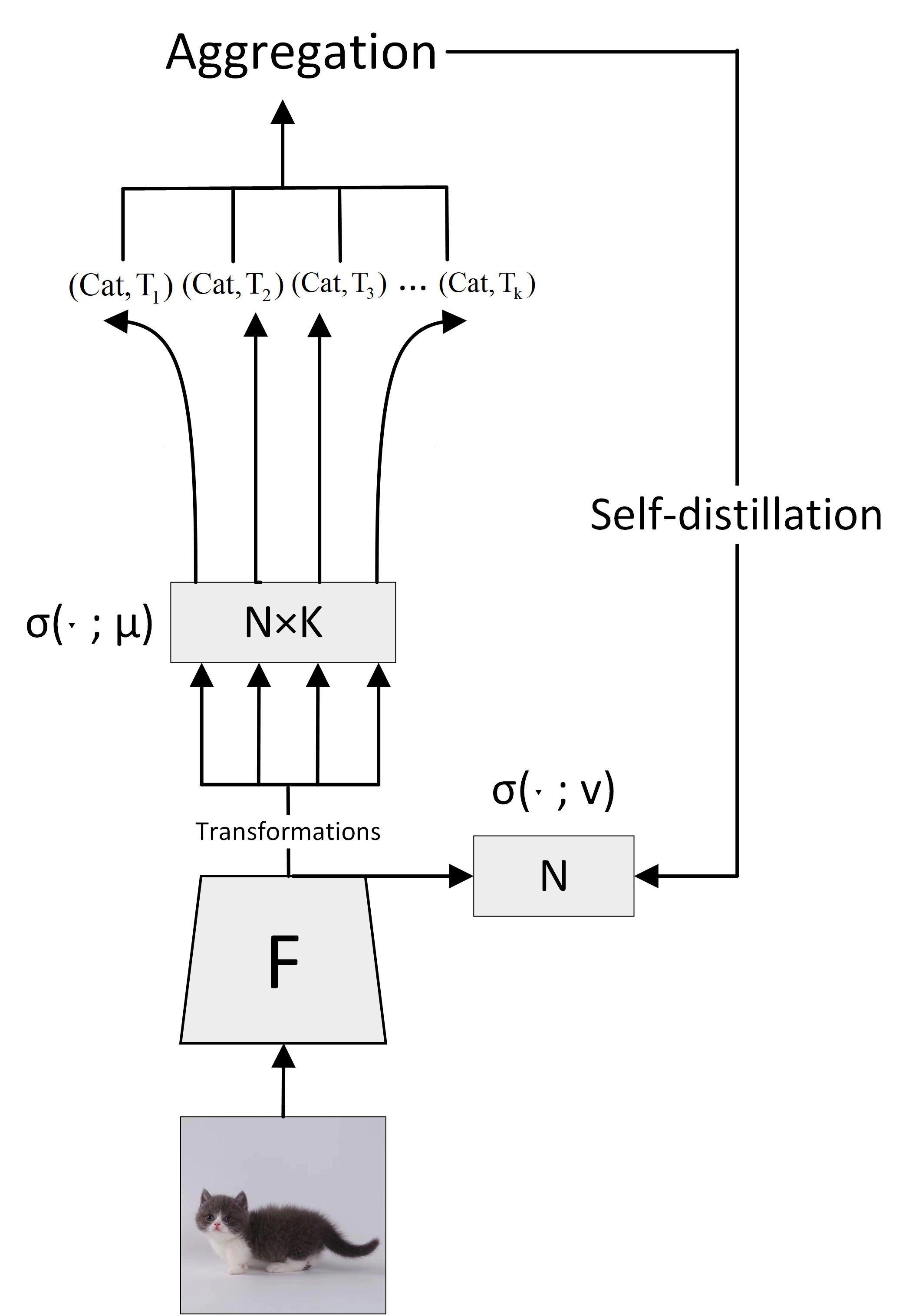}
	\caption{Illustrations of the single inference, the aggregated inference and the inference of self-distillation. If there are two joint classifiers in the model, we use the probability of all $ 2 \times K $ feature maps for aggregation.}
	\label{evaluation structure}
\end{figure}

In order to restore the basic network structure in the test phase, we introduce self-distillation \cite{hinton2015distilling} to transfer the knowledge of the joint classifier $\sigma(\cdot;\mu)$ to another single classifier $\sigma(\cdot;v)$ (see Figure \ref{evaluation structure}). In this way, we can only use the single classifier for inference without aggregation after training, i.e., images can be classified using only a basic network model.

\end{document}